\documentclass[10pt,twocolumn,letterpaper]{article}

\usepackage[pagenumbers]{cvpr} 

\usepackage{graphicx}
\usepackage{threeparttable}
\usepackage{amsmath}
\usepackage{amssymb}
\usepackage{booktabs}
\usepackage{multirow}
\usepackage{enumitem}

\usepackage[accsupp]{axessibility}

\newcommand{\tablestyle}[2]{\setlength{\tabcolsep}{#1}\renewcommand{\arraystretch}{#2}\centering\footnotesize}
\newlength\savewidth
\newcommand{\myplus}[1]{\color{green}{\tiny{}}}
\newcommand{\myminus}[1]{\color{red}{\tiny{}}}

\newcommand\mypara[1]{\vspace{1mm}\noindent\textbf{#1}}

\usepackage[pagebackref,breaklinks,colorlinks]{hyperref}

\usepackage[capitalize]{cleveref}
\crefname{section}{Sec.}{Secs.}
\Crefname{section}{Section}{Sections}
\Crefname{table}{Table}{Tables}
\crefname{table}{Tab.}{Tabs.}

\def\cvprPaperID{7432} 
\def\confName{CVPR}
\def\confYear{2022}

\usepackage{multicol}
\begin{document}

\title{Knowledge Distillation via the Target-aware Transformer}

\author{Sihao Lin$^{1,3}$\footnotemark[2]\, \footnotemark[3]\;,\;
Hongwei Xie$^{2}$\footnotemark[2]\;,\;
Bing Wang$^{2}$,\;
Kaicheng Yu$^{2}$,\\
Xiaojun Chang$^{3}$\footnotemark[4]\;,\;
Xiaodan Liang$^{4}$,\;
Gang Wang$^{2}$\\
$^1$RMIT University\;
$^2$Alibaba Group\;
$^3$ReLER, AAII, UTS\;
$^4$Sun Yat-sen University\\
{\tt\small \{linsihao6, hongwei.xie.90, Kaicheng.yu.yt, xdliang328\}@gmail.com}\\
{\tt\small \{fengquan.wb, wg134231\}@alibaba-inc.com,}\;
{\tt\small xiaojun.chang@uts.edu.au}
}
\maketitle

\begin{abstract}

Knowledge distillation becomes a de facto standard to improve the performance of small neural networks. Most of the previous works propose to regress the representational features from the teacher to the student in a one-to-one spatial matching fashion. However, people tend to overlook the fact that, due to the architecture differences, the semantic information on the same spatial location usually vary. This greatly undermines the underlying assumption of the one-to-one distillation approach. To this end, we propose a novel one-to-all spatial matching knowledge distillation approach. Specifically, we allow each pixel of the teacher feature to be distilled to all spatial locations of the student features given its similarity, which is generated from a target-aware transformer. Our approach surpasses the state-of-the-art methods by a significant margin on various computer vision benchmarks, such as ImageNet, Pascal VOC and COCOStuff10k. Code is available at \url{https://github.com/sihaoevery/TaT}.

\end{abstract}
\footnotetext[4]{Corresponding Author.}
\footnotetext[2]{Equal contribution.}
\footnotetext[3]{Part of the work done when as an intern in DAMO Academy.}

\vspace{-5mm}
\section{Introduction}
\label{sec:intro}
Knowledge distillation~\cite{Hinton2015DistillingTK,MishraM18} refers to a simple technique to improve the performance of any machine learning algorithms. One common scenario is to distill the knowledge from a larger teacher neural network to a smaller student one, such that the performance of student model can be significantly boosted comparing to training the student model alone. Concretely, people formulate an external loss function that guides the student feature map to mimic teacher's. Recently, it has been applied to various downstream applications, such as model compression~\cite{Yim2017AGF,Tian2020ContrastiveRD}, continual learning~\cite{LiH18a}, and semi-supervised learning \cite{chen2020big}.

\begin{figure}[!t]
    \centering
    \includegraphics[scale=0.5]{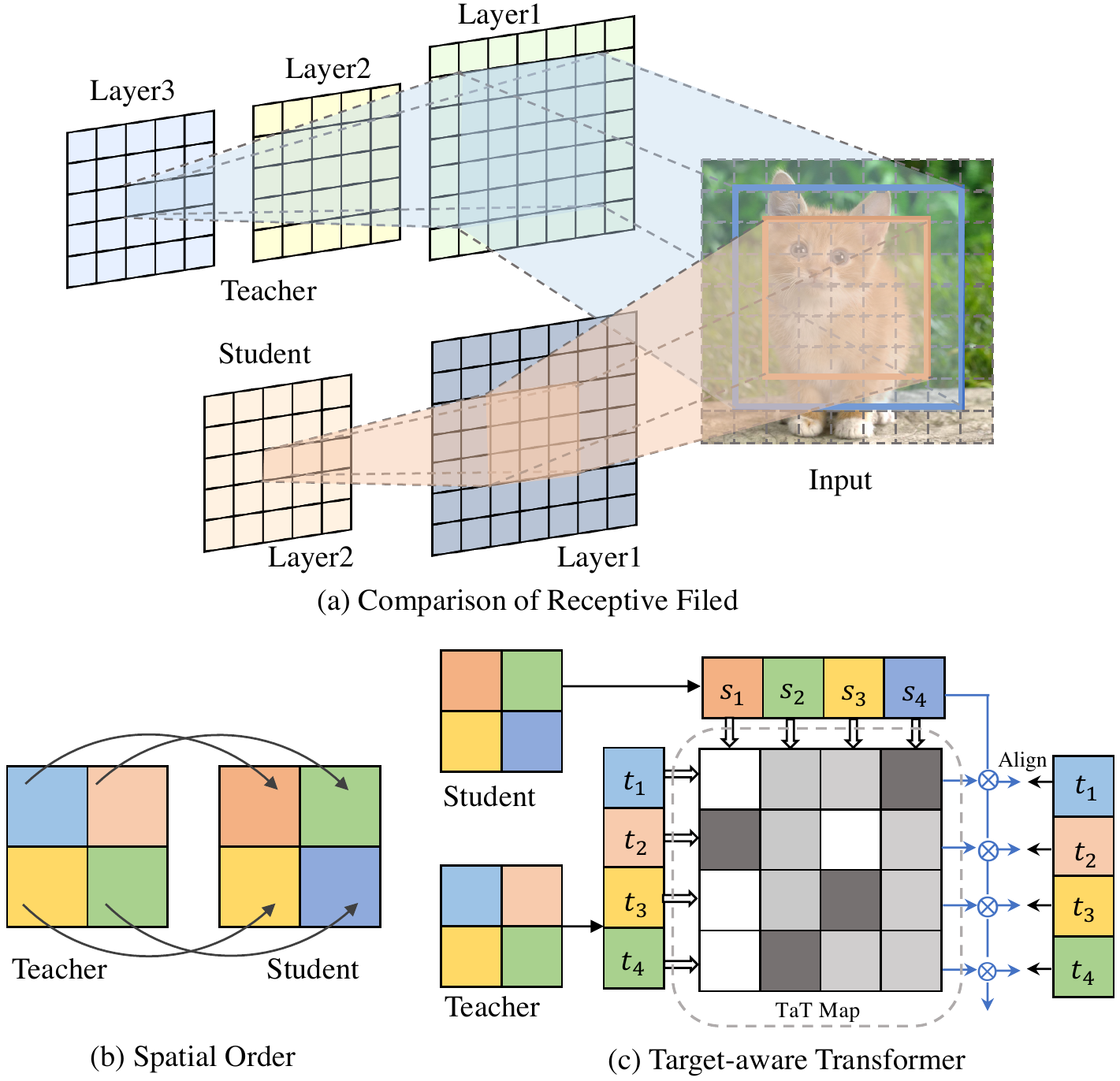}
    \caption{\textbf{Illustration of semantic mismatch.} Suppose that teacher and student are the 3-layers and 2-layers convnets with kernel size $3\times 3$ and stride $1\times 1$. (a) shows the receptive field of the middle pixel of the final feature map, where the blue box represents the teacher's receptive field and the orange box is that of the student's. Since teacher model has more convolutional operations, the resulting teacher feature map has a larger receptive field and thus contains richer semantic information. (b) Hence, directly regressing the student's and teacher's feature in a one-to-one spatial matching fashion may be suboptimal. 
    (c) We proposed a one-to-all knowledge distillation via a target-aware transformer that can let the teacher's spatial components be distilled to the entire student feature maps. 
    }
    \label{fig:motivation}
    \vspace{-0.8cm}
\end{figure}

Earlier works only distill the knowledge from the final layer of neural networks, for example, the ``logits'' in image classification task~\cite{Hinton2015DistillingTK,ba2013deep}. Recently, people discover that distilling the intermediate feature maps is a more effective approach to boost the student's performance. This line of works encourage similar patterns to be elicited in the spatial dimensions~\cite{Romero2015FitNetsHF,Zagoruyko2017PayingMA},
and is constituted as state-of-the-art knowledge distillation approach~\cite{Ji2021ShowAA,chen2021distilling}.

To compute the distillation loss of the aforementioned approach, one need to select the source feature map from the teacher and the target feature map from the student, where these two feature maps must have the same spatial dimension. As shown in Figure~\ref{fig:motivation}~(b), the loss is computed in a one-to-one spatial matching fashion, that is formulated as a summation of the distance between the source and the target features at each spatial location. One underlying assumption of this approach is the spatial information of each pixel is the same. In practice, this assumption is commonly not valid due to the fact that student model usually has fewer convolutional layers than the teacher. One example is shown in Figure~\ref{fig:motivation}~(a), even at the same spatial location, the receptive field of student feature is often significantly smaller than the teacher's and thus contains less semantic information. 
In addition, recent works~\cite{chen2017rethinking,dai2017deformable,Zagoruyko2016WideRN,szegedy2015going} evidences the importance of receptive field's influence on the model representation power. Such discrepancy is a potential reason that the current one-to-one matching distillation leads to sub-optimal results. 

To this end, we propose a novel one-to-all spatial matching knowledge distillation approach. In Figure~\ref{fig:motivation}~(c), our method distills the teacher's features at each spatial location into all components of the student features through a parametric correlation, \ie., the distillation loss is a weighted summation of all student components. To model such correlation, we formulate a transformer structure that reconstructs the corresponding individual component of the student features and produces an alignment with the target teacher feature. We dubbed this target-aware transformer. As such, we use parametric correlations to measure the semantic distance conditioned on the representational components of student feature and teacher feature to control the intensity of feature aggregation, which address the downside of one-to-one matching knowledge distillation.

As our method computes the correlation between feature spatial locations, it might become intractable when feature maps are large. To this end, we extend our pipeline in a two-step hierarchical fashion:
1) instead of computing correlation of all spatial locations, we split the feature maps into several groups of patches, then performs the one-to-all distillation within each group; 2) we further average the features within a patch into a single vector to distill the knowledge. This reduces the complexity of our approach by order of magnitudes.

We evaluate the effectiveness of our method on two popular computer vision tasks, image classification and semantic segmentation. 
On the ImageNet classification dataset, the tiny ResNet18 student can be boosted from 70.04\% to 72.41\% in terms of the top-1 accuracy, and surpasses the state-of-the-art knowledge distillation by 0.8\%. As for the segmentation task on COCOStuff10k, comparing to the previous approaches, our approach is able to boost the compact MobilenetV2 architecture by 1.75\% in terms of the mean intersection of union~(mIoU).

\vspace{1mm}
Our contributions can be summarized as follows:
\begin{itemize}[leftmargin=*,itemsep=0pt,topsep=0pt]
    \item We propose the knowledge distillation via a target-aware transformer, which enables the whole student to mimic each spatial component of the teacher respectively. In this way, we can increase the matching capability and subsequently improve the knowledge distillation performance.
    \item We propose the hierarchical distillation to transfer local features along with global dependency instead of the original feature maps. This allows us to apply the proposed method to applications, which are suffered from heavy computational burden because of the large size of feature maps.
    \item We achieve state-of-the-art performance compared against related alternatives on multiple computer vision tasks by applying our distillation framework.
\end{itemize}

\begin{figure*}[t!]
  \centering
  \includegraphics[width=\textwidth]{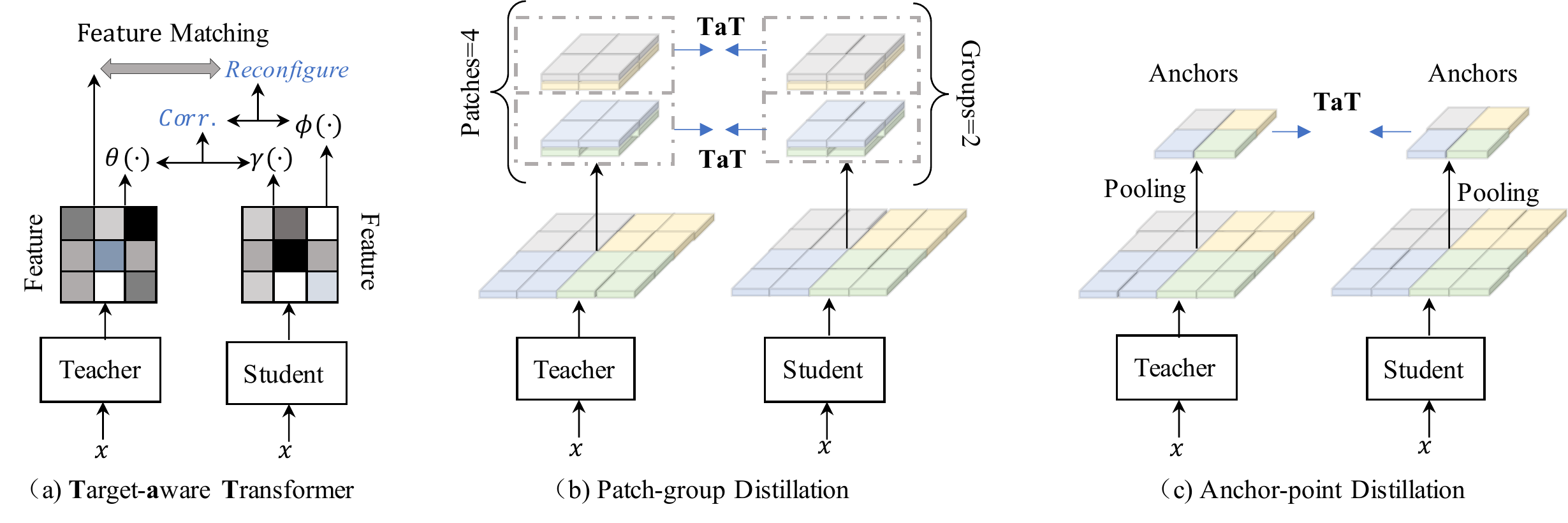}
  \vspace{-5mm}
  \caption{Illustration of our framework. (a) \textbf{Target-aware Transformer}. Conditioned on the teacher feature and the student feature, the transformation map \textit{Corr.} is computed and then applied on the student feature to reconfigure itself, which is then asked to minimize the L$_2$ loss with the corresponding teacher feature. (b) \textbf{Patch-group Distillation}. Both teacher and student features are to be sliced and rearranged as groups for distillation. By concatenating the patches within a group, we explicitly introduce the spatial correlation among the patches beyond the patches themselves. (c) \textbf{Anchor-point Distillation}. Each color indicates a region. We use average pooling to extract the \textit{anchor} within a local area of the given feature map, forming the new feature map of a smaller size. The generated anchor-point features will participate in the distillation.}
  \label{fig:framework}
  \vspace{-5mm}
\end{figure*}

\section{Related Works}
The seminal work~\cite{Hinton2015DistillingTK} introduced the idea of knowledge distillation. Specifically, Hinton \etal proposed to distill the logits (before sotfmax layer) from teacher to student by minimizing the KL divergence, where a temperature factor is applied to soften the logits.
Since feature map contains richer representation, Romero \etal~\cite{Romero2015FitNetsHF} introduced the intermediate layer transfer between teacher and student. 
Lately, AT~\cite{Zagoruyko2017PayingMA} proposed several statistical methods to highlight the dominating area of the feature map and discarded low-response area as noise. Chen \etal~\cite{Chen2020CrossLayerDW} proposed the semantic calibration which allowed the student to learn from the most semantic-related teacher layer. In \cite{Ji2021ShowAA}, the feature similarities between teacher and student were calculated and then were used as weights to balance the feature matching. These early methods intuitively established the links between knowledge source (teacher) and distillation terminal (student) in the one-to-one manner by spatial order. 

However, they overestimated the prior of spatial order while neglected the issues of semantic mismatch, \ie, the pixels of teacher feature map often contains richer semantic compared to that of student on the same spatial location. We found that some works~\cite{Park2019RelationalKD, Passalis2018LearningDR,Peng2019CorrelationCF,Tung2019SimilarityPreservingKD,Yim2017AGF,huang2017like,Liu2021ICKD}, though unintended, have been proposed to relax the spatial constrain during feature transfer. Typically, they defined the relational graph, and similarity matrix in the feature space of teacher network and transferred it to the student network. For instances, Tung and Mori~\cite{Tung2019SimilarityPreservingKD} calculated the similarity matrix where each entry encoded the similarity between two instances. Liu \etal~\cite{Liu2021ICKD} measured the correlation between channels by inner-product. They condensed and compressed the entire feature to some properties (often scalar) and thus collapsed the spatial information. On the other hand, such process damaged the original teacher feature and may lead to sub-optimal solution. 
The spread of KD has also driven some methods designed for specific vision tasks including video captioning~\cite{pan2020spatio}, action recognition~\cite{wang2019progressive,cui2020knowledge}, object detection~\cite{chen2017learning,zhang2020improve,dai2021general} and semantic segmentation~\cite{liu2019structured,he2019knowledge,Wang2020IntraclassFV}. Regarding the semantic segmentation, these methods are indeed related to relation knowledge distillation which computes similarity matrix~\cite{Tung2019SimilarityPreservingKD}. To investigate the potential of our method, we also adapt the method to semantic segmentation with hierarchical distillation.

The success of Transformer \cite{vaswani2017attention} in NLP has attracted lots of attention from the community of computer vision \cite{dosovitskiy2020image,li2021ffa,liu2021swin,wang2021pyramid}. While the original ViT \cite{dosovitskiy2020image} suffered from computation burden, Liu \etal \cite{liu2021swin} proposed the shifted-window that computes the attention on patch-level. The Pyramid ViT \cite{wang2021pyramid} proposed a progressive shrinking pyramid that adjusts the scale of feature map.
\section{Method}
\label{sec:method}

In this section, we first briefly describe the fundamental elements of feature map knowledge distillation and then introduce the general formulation of our knowledge distillation via a target-aware transformer. As our method computes the point-wise correlation of the given feature maps, the computational complexity becomes intractable on large-scale features, we then introduce the hierarchical distillation approach to address this limitation.

\subsection{Formulation}
\label{sec:formulation}
Suppose the teacher and the student are two convolutional neural networks, denoted by $T$ and $S$.
$F^T\in {\mathbb{R}^{H\times W\times C}}$ and ${F^S}\in \mathbb{R}^{H\times W\times C^{'}}$ denote the teacher feature and student feature respectively, where $H$ and $W$ are the height and width of the feature map, and $C$ represents the channel numbers. In the pioneer work~\cite{Hinton2015DistillingTK}, the distillation loss is formulated by a distance of features that come from the last layer of the networks. For example, in the image classification domain, it refers to the  ``logits'' before going in the softmax layer and cross-entropy loss. Concretely, the vanilla distillation loss is defined as:
\begin{equation}
    \mathcal{L}_{\rm{KL}}={\rm{KLD}}(\sigma(\frac{T(x)}{\tau}),\sigma(\frac{S(x)}{\tau})),
\end{equation}
\noindent where $\rm{KLD}(\cdot)$ measures the Kullback-Leibler divergence, $\sigma(\cdot)$ is the softmax function, $T(x)$ and $S(x)$ are the output logits given specific input $x$, and $\tau$ is the temperature factor. 
Without loss of generality, we assume that $C^{'}$ aligns with $C$ and reshape both $F^T$ and $F^S$ into 2D matrices:
\begin{equation}
\begin{aligned}
    f^s={\rm{\Gamma}}(F^S)\in \mathbb{R}^{N \times C},\\ 
    f^t={\rm{\Gamma}}(F^T)\in \mathbb{R}^{N \times C}.
\end{aligned}
\end{equation}
\noindent Here $\rm{\Gamma}(\cdot)$ is a function that flattens the 3D feature tensor into the 2D matrix where each row of the matrix is associated with a pixel in the feature tensor by spatial order and $N=H\times W$. We can describe $f^s$ and $f^t$ as two sets of the pixels with cardinality $N$ :
\begin{equation}
\begin{aligned}
    {f^s}^\top=[f^s_1,f^s_2,f^s_3,\dots,f^s_N],\\ 
    {f^t}^\top=[f^t_1,f^t_2,f^t_3,\dots,f^t_N].
\end{aligned}
\end{equation}

Previous work \cite{Romero2015FitNetsHF} simply minimize the discrepancy between two sets $f^s$ and $f^t$ in a one-to-one spatial matching manner, we denote this approach feature matching~(FM):
\begin{equation}
\label{eq:FM}
    \mathcal{L}_{\rm{FM}} = ||F^S-F^T||_2 = \sum_{i=1}^N ||f^s_i-f^t_i||_2.
\end{equation}
\noindent This formulation assumes that the semantic distributions of the teacher and the student match exactly. 
However, as mentioned earlier, for the feature maps of the teacher network, which usually encompasses more layers and larger feature channels, the spatial information of the same pixel location contains a richer semantic information compare to the student network. Directly regressing the features in a pixel-wise manner may lead to suboptimal distillation results. 

To this end, we propose a one-to-all spatial matching knowledge distillation pipeline that allows the each feature location of the teacher to teach the entire student features in a dynamic manner.
To make the whole student mimic a spatial component of the teacher, we propose the \textbf{T}arget-\textbf{a}ware \textbf{T}ransformer (\textbf{TaT}) to pixel-wisely reconfigure the semantic of student feature in the certain position. Given a spatial component (alignment target) of the teacher, we use \textbf{TaT} to guide the whole student to reconstruct the feature in its corresponding location. Conditioned on the alignment target,  \textbf{TaT} should reflect the semantic similarity with the components of the student feature. We use a linear operator to avoid changing the distribution of student semantics. The formulation of transformation operator $W^i$ can be defined as:
\begin{equation}
\label{eq:spe}
\begin{aligned}
   W^i&= \sigma(\langle {f^s_1},{f^t_i}\rangle,\langle {f^s_2},{f^t_i}\rangle,\dots,\langle {f^s_N},{f^t_i}\rangle)\\
   &=[{w^i_1},{w^i_2},\dots,{w^i_N}],
\end{aligned}
\end{equation}
\noindent where $f^t_i$ and $f^s_i$ denote the corresponding $i$-th components of teacher and student, $\langle \cdot,\cdot \rangle$ represents the inner-product and  $\|W^{i}\|=1$. We use inner-product to measure the semantic distance and softmax function for normalization. 
Each entry of $W^{i}$ is like the gate and controls the amount of semantic that will be propagated to the $i$-th reconfigured point. By aggregating the these related semantic across all the components, we have the result:
\begin{equation}
\label{eq:self-essem}
\begin{aligned}
   {f^s_i}^{'}={w^i_1}\times{f^s_1}+{w^i_2}\times{f^s_2}+\dots+{w^i_N}\times{f^s_N}.
\end{aligned}
\end{equation}
\noindent The Eq. \ref{eq:spe} and Eq. \ref{eq:self-essem} can be combined and rewritten as the form of matrix multiplication: ${f^s_i}^{'}=\sigma(f^{s}\cdot {f^t_i})\cdot f^{s}$. 

Note this is the simple non-parametric method that only depends on the original features. To facilitate the training, we introduce the parametric method with the extra linear transformation applied on the student feature and teacher feature. We observe that parametric version performs better than non-parametric one in ablation study. Guided by the target-aware transformer, the reconfigured student feature can be formulated as:
\begin{equation}
\label{eq:mul-self-essem}
    {f^s}^{'}=\sigma(\gamma(f^{s})\cdot \theta({f^t})^{\top})\cdot \phi(f^{s}),
\end{equation}
\noindent where $\theta(\cdot)$, $\gamma(\cdot)$ and $\phi(\cdot)$ are the linear functions consisting of $3\times 3$ conv layer plus the BN layer \cite{ioffe2015batch}. We compare the parametric \textbf{TaT} to non-parametric one to analyse the effectiveness brought by these linear functions in the Section~\ref{sec:ablation}. 
In the case that the channel numbers of $F^S$ do not match with that of $F^T$, $\gamma(\cdot)$ can help with alignment.

After reconfiguration, each component of ${f^s}^{'}$ aggregates the meaningful semantic from the original feature, which enhances the expressivity. We do not require the student to reconstruct the teacher feature in a pixel-to-pixel manner. Indeed, our model allows the student to act as a whole to mimic the teacher. The resulting ${f^s}^{'}$ is lately asked to minimize the L$_2$ loss with the teacher feature. The objective for \textbf{TaT} knowledge distillation can be given by:
\begin{equation}
    \mathcal{L}_{\rm{TaT}}= ||{f^s}^{'}-f^t||_2.
    \label{eq:fm}
\end{equation}

Finally, the total loss of our proposed method can be defined by: 

\begin{equation}
\label{eq:objective}
    \mathcal{L}=\alpha\mathcal{L}_{\rm{Task}}+\beta\mathcal{L}_{\rm{KL}}+\epsilon\mathcal{L}_{\rm{TaT}},
\end{equation}
\noindent Here $\mathcal{L}_{\rm{Task}}$ can be any loss on the generic machine learning tasks. $\alpha$, $\beta$ and $\epsilon$ are the weight factors to balance the loss. 
Empirically, we find that our model benefits from $\mathcal{L}_{\rm{KL}}$. However, the model can achieve state-of-the-art without the help of $\mathcal{L}_{\rm{KL}}$.


\subsection{Hierarchical Distillation}
\label{sec:hierarchical}
The proposed \textbf{TaT} lift the limitation of previous one-to-one spatial matching fashion. 
However, the computation complexity of \textbf{TaT} map will become intractable
when it comes to a large feature map. Assuming the spatial dimensions of the feature map are $H$ and $W$, this means the computation complexity will reach $\mathcal{O}(H^2\cdot W^2)$. 
Therefore, we propose a hierarchical distillation approach to address this large feature map limitation. 
It contains two steps: 1) patch-group distillation that splits the entire feature maps into smaller patches, so to distill local information from the teacher to the student; 2) we further summarize the local patches into one vector and distill this for global information.
\subsubsection{Patch-group Distillation}
\label{sec:seq}

As mentioned above, as the spatial dimension of input feature maps increases, distillation becomes more difficult. A straightforward solution~\cite{Liu2021ICKD} is to divide the feature map into patches and perform distillation within patches individually. 
However, the correlation between patches is completely ignored, resulting in sub-optimal solutions.

In contrast to Liu \etal~\cite{Liu2021ICKD}, we propose the patch-group distillation (See Figure \ref{fig:framework} (b)) that allows the student to learn the local feature from patches and retain the correlation among them to some extent. Given the original student feature $F^S$ and teacher feature $F^T$, they are partitioned into $n\times m$ patches of size $h\times w$, where $h=H/n$, $w=W/m$. They are further arranged as $g$ groups sequentially where each group contains $p=n\cdot m/g$ patches. Specifically, the patches in a group will be concatenated channel-wisely, forming a new tensor of size $h\times w \times c\cdot g$ that would be used for distillation lately. In this way, each pixel of the new tensor contains the features from $p$ positions of the original feature, which explicitly includes the spatial pattern. Therefore, during the distillation, the student can learn not only the single pixel but the correlation among them. Intuitively, a larger group will introduce richer correlation but complex correlation will turn to difficult to learn. We study the effectiveness of different group sizes in the experiments.

Similar to the formulation presented in Section \ref{sec:formulation}, the patch-group distillation can be given by simply replacing the original input with the reorganized one, and the variant is denoted as $\mathcal{L}_{\rm{TaT}}^\mathcal{P}$. To relax the strict constraints of the spatial pattern in the patch-group, we set the $\theta(\cdot)$ as linear transformation in our experiments.
\subsubsection{Anchor-point Distillation}
\label{sec:anchor}
The patch-group distillation can learn the fined-grained feature on the patch level and retain the spatial correlation among the patches to some extent. However, it is not capable of perceiving the long-range dependency. As will see in the ablation study, the attempt to preserve the global correlation through concatenating all the patches would fail. 
For complex scenes, long-range dependency is important to capture the relation (\eg layout) of different objects.

We address the conundrum by the proposed anchor-point distillation. As shown in Figure~\ref{fig:framework} (c), we summarize the local area to compact representation, referred to \textit{anchor}, within a local area that is representative to describe the semantic of the given area, forming the new feature map of smaller size. Since the new feature map consists of the summary of the original feature, it can approximately substitute the original one to obtain the global dependency. We simply use average pooling to extract the anchor points. Then all the anchors are scattered back to the associated position to form a new feature map. The anchor-point feature is used for distillation as described in Section \ref{sec:formulation} and the objective is denoted as $\mathcal{L}_{\rm{TaT}}^\mathcal{A}$. The patch-group distillation enables the student to mimic the local feature while the anchor-point distillation allows the student to learn the global representation over the coarse anchor-point feature, which are complementary to each other. Therefore, the combination of these two objectives can bring the best of two worlds. Our objective designed for semantic segmentation can be written by:
\begin{equation}
    \mathcal{L}_{\rm{Seg}}=\alpha\mathcal{L}_{\rm{CE}}+
    \delta\mathcal{L}_{\rm{TaT}}^\mathcal{P}+
    \zeta\mathcal{L}_{\rm{TaT}}^\mathcal{A}
    \label{eq:seg}
\end{equation}


\begin{table*}[ht]
\centering
\caption{\textbf{Top-1 accuracy(\%) on Cifar-100.} The loss term $\mathcal{L}_{\rm{KL}}$ in Eq.~\ref{eq:objective} is removed in this experiment.
}
\vspace{-0.2cm}
\resizebox{2.0\columnwidth}{!}{
\begin{tabular}{l|ccccccc} 
\toprule
\multirow{3}{*}{Method} & \multicolumn{7}{c}{Network Architecture}  \\ 
\cmidrule{2-8}
& WRN-40-2 & WRN-40-2  & ResNet56 & ResNet110 & ResNet110 & ResNet32$\times$4 & VGG13 \\
& WRN-16-2 & WRN-40-1  & ResNet20 & ResNet20  & ResNet32  & ResNet8$\times$4  & VGG8 \\ \midrule
Teacher &75.61 &75.61 &72.34 &74.31 &74.31 &79.42 &74.64  \\ 
Vanilla &73.26 &71.98 &69.06 &69.06 &71.14 &72.50 &70.36  \\  
KD \cite{Hinton2015DistillingTK} &74.92 &73.54 &70.66 &70.67 &73.08 &73.33 &72.98  \\
FitNet \cite{Romero2015FitNetsHF} &73.58\myminus{1.34}&72.24\myminus{1.30}&69.21\myminus{1.45}&68.99\myminus{1.68}&71.06\myminus{2.02}&73.50\myplus{0.17}&71.02\myminus{1.96} \\ 
AT \cite{Zagoruyko2017PayingMA}&74.08\myminus{0.84}&72.77\myminus{0.77}&70.55\myminus{0.11}&70.22\myminus{0.45}&72.31\myminus{0.77}&73.44\myplus{0.11}&71.43\myminus{1.55} \\ 
SP \cite{Tung2019SimilarityPreservingKD} &73.83\myminus{1.09}&72.43\myminus{1.11}&69.67\myminus{0.99}&70.04\myminus{0.63}&72.69\myminus{0.39}&72.94\myminus{0.39}&72.68\myminus{0.20} \\ 
CC \cite{Peng2019CorrelationCF}&73.56\myminus{1.36}&72.21\myminus{1.33}&69.63\myminus{1.03}&69.48\myminus{1.19}&71.48\myminus{1.60}&72.97\myminus{0.36}&70.71\myminus{2.27} \\ 
RKD \cite{Park2019RelationalKD}&73.35\myminus{1.57}&72.22\myminus{1.32}&69.61\myminus{1.05}&69.25\myminus{1.42}&71.82\myminus{1.26}&71.90\myminus{1.43}&71.48\myminus{1.50} \\
PKT \cite{Passalis2018LearningDR}&74.54\myminus{0.38}&73.45\myminus{0.09}&70.34\myminus{0.32}&70.25\myminus{0.42}&72.61\myminus{0.47}&73.64\myplus{0.31}&72.88\myminus{0.10} \\ 
FSP \cite{Yim2017AGF}&72.91\myminus{2.01}&NA&69.95\myminus{0.71}&70.11\myminus{0.56}&71.89\myminus{1.19}&72.62\myminus{0.71}&70.20\myminus{2.78}\\
NST \cite{huang2017like}&73.68\myminus{1.24}&72.24\myminus{1.30}&69.60\myminus{1.06}&69.53\myminus{1.14}&71.96\myminus{1.12}&73.30\myminus{0.03}&71.53\myminus{1.45}\\ 
CRD \cite{Tian2020ContrastiveRD}&75.48\myplus{0.56}&74.14\myplus{0.60}&71.16\myplus{0.50}&71.46\myplus{0.79}&73.48\myplus{0.40}&75.51\myplus{2.18}&73.94\myplus{0.96}\\ 
ICKD~\cite{Liu2021ICKD} &75.64 &74.33 &\textbf{71.76} &71.68 &73.89 &75.25 &73.42 \\
\midrule
Ours w/o $\mathcal{L}_{\rm{KL}}$&\textbf{76.06}\myplus{1.10}&\textbf{74.97}\myplus{1.43}&71.59\myplus{0.73}&\textbf{71.70}\myplus{1.03}&\textbf{74.05}\myplus{0.88}&\textbf{75.89}\myplus{2.17}&\textbf{74.39}\myplus{1.41}\\
\bottomrule
\end{tabular}}
\label{tab:cifar_sota}
\vspace{-4mm}
\end{table*}
\section{Experiment}
\label{sec:experiment}
In this section, we empirically evaluate the effectiveness of the proposed method through extensive experiments.
On image classification, we leverage the commonly used benchmark in knowledge distillation such as
Cifar-100\cite{krizhevsky2009learning} and ImageNet \cite{deng2009imagenet}, and show our model can improve the student performance by a significant margin compared to many state-of-the-art baselines.
In addition, we extend our method to another popular computer vision task, semantic segmentation to further demonstrate the generalization ability of our method. We nonetheless provide a detailed ablation study in the end of this section.

\subsection{Datasets}
\mypara{Cifar-100}~\cite{krizhevsky2009learning}. This benchmark contains 100 categories including 600 samples each. For each category, there are 500 images for training while 100 images for testing. 
We report top-1 accuracy as evaluation metric.   

\mypara{ImageNet}~\cite{deng2009imagenet}. This is a challenging benchmark for image classification including more than one million training samples with 1,000 categories. Similarly, we report the top-1 accuracy to measure the model performances. 

\mypara{Pascal VOC}~\cite{Everingham10}.
This benchmark contains 20 foreground classes with a background class. It provides 1,464 training, 1,499 validation, and 1,456 testing samples. Apart from the fine annotated samples, we also use additional coarse annotated images from \cite{Hariharan2011SemanticCF} for training, resulting in 10,582 training samples. We report the mean Intersection over Union (mIoU) on the validation set to measure the proposed method.

\mypara{COCOStuff10k}~\cite{caesar2018coco}. The challenging dataset is developed on MSCOCO \cite{lin2014microsoft} by adding dense pixel-wise stuff label, resulting in 172 classes: 80 for thing, 91 for stuff, and 1 for unlabeled. It contains 9k training samples and 1k validation samples. We report the mIoU to evaluate our method.

\subsection{Implementation Details}
\mypara{Image classification.}
For the experiments on Cifar-100, we use SGD optimizer \cite{sutskever2013importance} and the total running epoch is set to 240. The initial learning rate is 0.05 with a decay rate 0.1 at epoch 150, 180, and 210. In terms of data augmentation, the input images will be randomly cropped and flipped horizontally. 
We use Bayesian optimization \cite{snoek2012practical} for hyper-parameters (\ie $\alpha$ and $\epsilon$ in Eq.~\ref{eq:objective}) searching. We report the exact values in the supplementary materials.
For the ImageNet experiments, we use the AdamW optimizer~\cite{loshchilov2017decoupled} and train all of the models for 100 epochs with a batch size of 2048. The initial learning rate is set to 1.6e-4 and decays by 0.1 at epoch 30, 60, and 90. We apply standard data augmentations including random crop and horizontal flip. We use a simple grid search on the hyper-parameters, and set $\alpha$=0.5, $\beta$=0.5 and $\epsilon$=0.1 in Eq.~\ref{eq:objective}.

\mypara{Semantic segmentation.}
We choose the DeepLabV3+~\cite{chen2018encoder} as the base architecture, where it contains a backbone to extract feature and a head to generate the segmentation results. For the teacher, we follow~\cite{chen2018encoder} to use the ResNet101 as the backbone model. For the student, we select two networks, ResNet18 which shares a similar architecture design as the backbone, and MobileNetV2\cite{sandler2018mobilenetv2} which is drastically different. 
We use random flip and Gaussian blur for data augmentation. The samples are randomly cropped and rescaled to $513\times 513$ during the training and are resized to the same resolution during the testing. The student backbone ResNet18 is trained for 100 epochs with an initial learning rate 7e-3 on the Pascal VOC and 1e-2 on COCOStuff10k respectively. For MobileNetV2, the learning rate is set to 7e-3 for all datasets. We incorporate the cosine learning rate scheduler for all experiments.  
On Pascal VOC, the weight factors of Eq. \ref{eq:seg} are $\alpha$ = 1, $\delta$ = 0.1, and $\zeta$ = 0.05. In terms of COCOStuff10k, the weight factors are $\alpha$ = 1, $\delta$=0.6, $\zeta$=0.6 for ResNet18, and are $\alpha$=1, $\delta$=0.4, $\zeta$ = 0.4 for MobileNetV2.


\begin{table*}[thb]
\centering
\caption{Top-1 Accuracy(\%) on ImageNet validation set. The ResNet34 is employed as the teacher backbone and the ResNet18 is selected as the student backbone. 
Our method can boost the performance of the tiny ResNet18 beyond 72\% and outperforms other methods without $\mathcal{L}_{\rm{KL}}$.
}
\vspace{-0.2cm}
\resizebox{\textwidth}{!}{

\begin{tabular}{@{}l|ccccccc|ccccc|c@{}}  
\toprule

Method & Vanilla &AT \cite{Zagoruyko2017PayingMA} &CRD \cite{Tian2020ContrastiveRD} & SAD \cite{Ji2021ShowAA} &ICKD \cite{Liu2021ICKD} &KR~\cite{chen2021distilling} &Ours &KD \cite{Hinton2015DistillingTK} & SCKD \cite{Chen2020CrossLayerDW} &CC \cite{Peng2019CorrelationCF} &RKD \cite{Park2019RelationalKD} &Ours &Teacher\\
\midrule
w/ $\mathcal{L}_{\rm KL}$ && &&&&&& $\checkmark$ &$\checkmark$&$\checkmark$&$\checkmark$&$\checkmark$& - \\
Top-1& 70.04 &70.59 &71.17 &71.38 &71.59 &71.61 &72.07 &70.68 &70.87 &70.74 &71.34 &\textbf{72.41} & 73.31\\

\bottomrule
\end{tabular}
}\label{tab:imagenet}
\vspace{-4mm}
\end{table*}
\vspace{-2mm}
\subsection{Image Classification}
\vspace{-2mm}
\mypara{Results on Cifar-100.} 
To show the generalization ability of our method, we applied our distillation approach to various network architectures, including ResNet~\cite{He2016DeepRL}, VGG~\cite{simonyan2014very} and WideResNet~\cite{Zagoruyko2016WideRN}. 
And in these experiments, we set $\theta(\cdot)$ as an identical function instead of linear transformation, \eg, Conv+BN.
As shown in Table~\ref{tab:cifar_sota}, our method surpasses all baselines on six out of seven teacher-student settings, often by a significant margin. This evidences the effectiveness and generalization ability of our approach. 
Compared to the closest baseline, FitNet\cite{Romero2015FitNetsHF}, which directly 
computes the distillation loss in one-to-one fashion, our approach improves on average 2.72\%. 
The results when distilling to ResNet20 is interesting. In this case, using a less powerful teacher, ResNet56, results a better student performance on average comparing to using ResNet110. In particular, directly distilling the feature in one-to-one fashion deteriorates the student's performance compared to vanilla training. 
Our distillation approach addressed such mismatch and achieves 71.70\% which is 2.64\% better than the baseline. We also compare our method to the comparison methods in the setting with extra  $\mathcal{L}_{\rm{KL}}$ in appendix.

\vspace{-1mm}
\mypara{Results on ImageNet.} 
Since Cifar-100 only contains 50,000 training images, we further evaluate our approach on a more challenging dataset. Here, we choose ResNet34 and ResNet18 as teacher and student model respectively. We show the Top-1 accuracy of the student and teacher model in Table~\ref{tab:imagenet}. Our method outperforms the state-of-the-art methods by a significant margin. Notice that, even without the help of $\mathcal{L}_{\rm{KL}}$, our model can reach 72.07\% on a tiny ResNet18, comparing to some methods which rely on the $\mathcal{L}_{\rm{KL}}$ by more than 1\%. 
When enabling $\mathcal{L}_{\rm{KL}}$, the proposed method can further improve the Top-1 accuracy of the student to 72.41\%. 
Compared to SCKD\cite{Chen2020CrossLayerDW} which uses an attention mechanism to re-allocate the most semantic-related teacher layers to the student, our method has a significant improvement.
That means even matching two layers of teacher and student with similar semantics, the student may not be able to catch up with the teacher in the pixel-to-pixel manner due to semantic mismatch.
In contrast, our method leverages a target-aware transformer to address the semantic mismatch in a more efficient manner.

\begin{table}[]
    \centering
    \caption{\textbf{Comparing the semantic segmentation results (in mIoU\%) of different methods on Pascal VOC.}
     We can observe that our method surpasses all previous baselines by a significant margin. Specifically, on the popular compact architecture MobilenetV2, our method improves the student by 5.39\% comparing to the stand-alone training, and by 1.06\% comparing to the state-of-the-art method ICKD.
    ${\dag}$ indicates reproducing by training 100  epochs, using the official released code.}
    \vspace{-2mm}
    \resizebox{0.9\columnwidth}{!}
    {
    \tablestyle{12pt}{0.8}
    \begin{tabular}{@{}l|cc@{}}
    \toprule
         &ResNet18 &MobilenetV2   \\
        \midrule
        Teacher &78.43 &78.43 \\
        Student &72.07 &68.46     \\
        KD~\cite{Hinton2015DistillingTK}&73.74 &71.73     \\
        AT~\cite{Zagoruyko2017PayingMA}&73.01 &71.39     \\
        FitNet~\cite{Romero2015FitNetsHF}&73.31 &69.23     \\
        Overhaul$^{\dag}$~\cite{Heo2019ACO}&73.98 &72.30     \\
        ICKD~\cite{Liu2021ICKD}    &75.01 &72.79     \\
        Ours    &\textbf{75.76} &\textbf{73.85} \\
    \bottomrule
    \end{tabular}}
    \label{tab:seg_voc}
    \vspace{-2mm}
\end{table}

\begin{table}[h]
    \centering
    \caption{Non-parametric vs. parametric implementation of target-aware transformer on ImageNet, where check mark indicates applying linear function.}
    \vspace{-3mm}
    \resizebox{0.8\columnwidth}{!}
    {
    \tablestyle{15pt}{0.9}
    \begin{tabular}{cc|c}
    \toprule
         $\theta(\cdot)$ &$\gamma(\cdot)$ &Top-1 Acc.  \\
         \midrule
         & &72.22\\
         &\checkmark&\textbf{72.41}\\
         \checkmark &\checkmark &72.35\\
         \bottomrule
    \end{tabular}
    }
    \label{tab:non-parametric_imagenet}
    \vspace{-2mm}
\end{table}


\begin{table}[h]
    \centering
    \caption{\textbf{Comparing the semantic segmentation results (in mIoU\%) of different methods on COCOStuff10k.}
    As most baselines do not provide the code on the COCO dataset except KR, we only compare our method to KR in this case. We reproduce the baseline using the official code with the same training procedure. Our method surpasses the baseline by nearly 2\%, and further demonstrates the effectiveness of our approach.
    }
    \vspace{-2mm}
    \begin{tabular}{@{}l|cccc@{}}
    \toprule
        &Sutdent   &KR~\cite{chen2021distilling}&Our &Teacher  \\
        \midrule
        ResNet18    &26.33 &26.73 &\textbf{28.75} &33.10 \\
        MobilenetV2 &26.29 &26.63 &\textbf{28.05}  &33.10 \\
    \bottomrule
    \end{tabular}
    \label{tab:seg_coco}
    \vspace{-6mm}
\end{table}
\vspace{-1mm}
\subsection{Semantic Segmentation}
\vspace{-1mm}
As the feature map size is fairly small when performing distillation on image classification, we plan to further investigate the generalization ability of our method on semantic segmentation, where the feature size is drastically larger. As in Section~\ref{sec:hierarchical}, we adapt our TaT method with the patch-group and anchor-point scheme. We select two popular benchmarks, Pascal VOC and COCOStuff10k, and present the results in Table \ref{tab:seg_voc} and Table \ref{tab:seg_coco} respectively. Our method clearly surpasses all baselines by a clear margin.
For instance, on Pascal VOC, the proposed model can improve the MobileNetV2 by more than 5\%, which shows great potential to unlock the hardware limitation. On the challenging benchmark COCOStuff10k, the model can improve the ResNet18 and MobileNetV2 by $2.42\%$ and $1.76\%$.

\begin{table}[t!]
\centering
\caption{\textbf{Impact of function $\theta(\cdot)$ on a variety of network architectures.} We report the top-1 accuracy on Cifar-100. $id$ indicates for identity mapping.  }
\vspace{-2mm}
\resizebox{0.85\columnwidth}{!}
{
\begin{tabular}{@{}ll|cc@{}}
    \toprule
    Teacher  &Student &Conv+BN & $id$  \\ 

    \midrule
    ResNet56 &ResNet20  &71.45 &\textbf{71.59} \\
    ResNet110 &ResNet20 &71.68 & \textbf{71.70} \\
    ResNet110 &ResNet32 &73.75 &\textbf{74.05}\\
    ResNet32$\times$4 &ResNet8$\times$4 &75.30 &7\textbf{5.89}\\
    VGG13 &VGG8 &73.48 &\textbf{74.39}\\
    \bottomrule
\end{tabular}
}
\label{tab:cifar_1x1}
\vspace{-3mm}
\end{table}

\begin{figure} [t]
    \centering 
    \includegraphics[width=.9\linewidth]{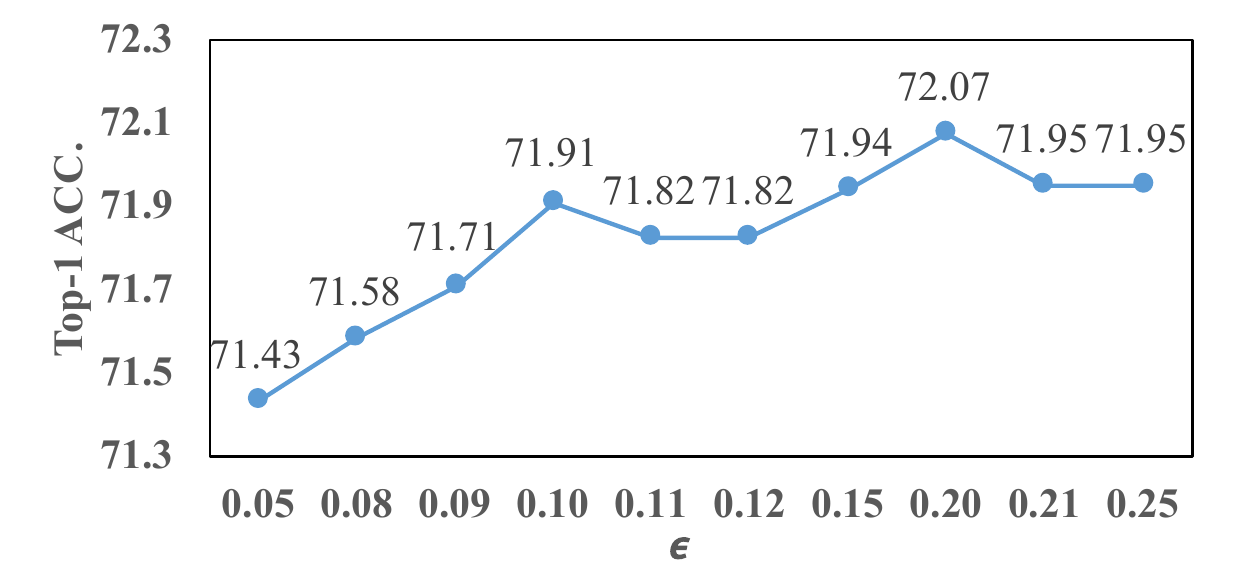}\vspace{-3mm}
    \caption{The performance of our model under different $\epsilon$ on ImageNet. Here the loss $\mathcal{L}_{\rm{KL}}$ is removed and $\alpha$ is set to 0.1.} 
    \label{fig:imgnet_sigma} 
    \vspace{-5mm}
\end{figure}

\subsection{Ablation Study}
\label{sec:ablation}
Here, we provide detailed ablation study to validate each component of our approach.

\mypara{Linear transformation functions.} 
We first study the impact of function $\theta(\cdot)$.
We are interested in that if the learning target (\ie teacher feature) is fixed, can the student adapt itself through target-aware transformer better? We compare different settings of $\theta(\cdot)$ including identical mapping against Conv$+$BN. The result on Cifar100 is presented on Table \ref{tab:cifar_1x1}. Surprisingly, the identical mapping for $\theta(\cdot)$ always performs better. 

We further investigate the non-parametric implementation by setting both $\theta(\cdot)$ and $\gamma(\cdot)$ as identical mapping on ImageNet (Table~\ref{tab:non-parametric_imagenet}). The result shows that the semi-parametric version performs best, where the fixed teacher and the linear transformation applying to student feature can facilitate the student to reconfigure itself.

\mypara{Validating $\epsilon$}. 
To investigate the efficacy brought by the proposed Eq. \ref{eq:fm}, we then further explore the different settings of the coefficient $\epsilon$ used in Eq. \ref{eq:objective} (See Figure~\ref{fig:imgnet_sigma}). When increased from 0.05 to 0.25, the objective $\mathcal{L}_{\rm{TaT}}$ can bring positive and stable effect.

\begin{table}[t!]
    \centering
    \caption{Contribution of patch-group and anchor-point distillation. We observe that patch-group distillation presents more efficacy. }
    \vspace{-2mm}
    \resizebox{0.8\columnwidth}{!}
    {\tablestyle{12pt}{0.9}
    \begin{tabular}{cc|c}
    \toprule
          Anchor-point&Patch-group& mIoU  \\
          \midrule
            &&72.07 \\ 
            \checkmark & &75.37 \\
            &\checkmark&75.63\\
            \checkmark & \checkmark&\textbf{75.76} \\
    \bottomrule
    \end{tabular}}
    \label{tab:seg_ablation}
    \vspace{-3mm}
\end{table}

\begin{table}[t!]
    \centering
    \caption{Performance (\%) and training time (minutes) of anchor-point distillation on Pascal VOC under different kernel sizes. }
    \vspace{-3mm}
    \resizebox{0.9\columnwidth}{!}
    {
    \tablestyle{7pt}{1.0}
    \begin{tabular}{@{}l|cccc@{}}
    \toprule
         Pooling kernel &$2\times 2$ &$4\times 4$&$8\times 8$&$16\times 16$ \\
        Training time &423 &403 &389 &374 \\
         mIoU       &\textbf{75.37} &75.27 &74.79 &74.56 \\
    \bottomrule
    \end{tabular}
    \label{tab:anchor}}
    \vspace{-5mm}
\end{table}

We also conduct the thorough experiments to understand the contribution brought by the proposed patch-group distillation $\mathcal{L}_{\rm{TaT}}^{\mathcal{P}}$ and anchor-point distillation $\mathcal{L}_{\rm{TaT}}^{\mathcal{A}}$. As discussed previously, $\mathcal{L}_{\rm{TaT}}^{\mathcal{A}}$ is proposed to learn the global representation to capture long-range dependency while $\mathcal{L}_{\rm{TaT}}^{\mathcal{P}}$ is designed to concentrate on local feature. 
By covering each one of them, the individual effectiveness of the two components can be examined.  
As shown in Table. \ref{tab:seg_ablation}, both objectives can improve the vanilla student significantly while $\mathcal{L}_{\rm{TaT}}^{\mathcal{P}}$ presents more efficacy. 
The combination of both components achieves the best performance, demonstrating that the two proposed objectives are complementary.

\mypara{Validating the anchor-point distillation.}
Then, we give more insight concerning the proposed objectives' functionality through sensitivity analysis. Specifically, we investigate the hyper-parameters that would influence the behavior of the training process. In terms of the anchor-point distillation, this work utilizes average pooling to extract the anchor in a local area from the original feature, forming the associated anchor-point feature. 
It is a trade-off between reducing computation overhead and summarizing fine-grained spatial information since a bigger kernel would reduce feature size along with more informative representation, \eg, when feature map size is reduced to $1 \times 1$, it degrades to ignoring the spatial information and posing one-to-one fashion distillation.
Thus we study the pooling kernel size that directly yields different feature resolutions.
The result exhibited in Table \ref{tab:anchor} shows that the amount of distillation calculation is greatly reduced with the increasing pooling size. On the other hand, excessive pooling range would omit useful and informative representation and damage the performance.
We also report the mean training time. All experiments are conducted on a single Nvidia A100 GPU (40GB memory) with Intel Xeon CPU (8 cores) for 3 times. 

\mypara{Validating the patch-group distillation.}
Next we analyze the two key factors of patch-group distillation, \ie patch size $h \times w$ and groups $g$. 
In Table \ref{tab:seq}, we found that generally, smaller patch size is advantageous to patch-group distillation and overlarge patch size, however, may be unfavourable since it approaches the original feature.  
Regarding the groups, it merges the patches as a group for joint distillation. In the experiment shown in Table \ref{tab:head}, the patch size is set to $8\times 8$, which divides the original feature map into $128/8*128/8=256$ patches. There are two extreme situations. When only one group is used, it indicates that all of the patches will be distilled jointly. On the contrary, using 256 groups means each patch is distilled individually. In this example, we found that 4 patches as a group can reach the best performance.

\begin{table}[t!]
    \centering
    \caption{Performance (\%) of patch-group distillation on Pascal VOC under different settings of patch size ($h \times w$). Groups is equal to patches $g=n \times m$.}
    \vspace{-3mm}
    \resizebox{1.0\columnwidth}{!}
    {\tablestyle{12pt}{1.0}
    \begin{tabular}{@{}l|cccc@{}}
    \toprule
         Patch size &32$\times$32 &16$\times$16 &8$\times$8 &4$\times$4  \\
         mIoU &75.33 &75.45 & \textbf{75.50} & 75.47\\
    \bottomrule
    \end{tabular}
    \label{tab:seq}}
    \vspace{-2mm}
\end{table}

\begin{table}[t!]
    \centering
    \caption{Performance (\%) of patch-group distillation on Pascal VOC under different settings of groups where patch size is $8\times 8$ and patch numbers is 256. }
    \vspace{-3mm}
    \resizebox{1.0\columnwidth}{!}
    {\tablestyle{10pt}{0.9}
    \begin{tabular}{@{}l|cccccc@{}} 
    \toprule
         Groups &1 &32 &64 &128 &256   \\
         mIoU &75.26&75.57&\textbf{75.63}&75.62&75.50\\
    \bottomrule
    \end{tabular}
    \label{tab:head}}
\vspace{-6mm}
\end{table}

\vspace{-2mm}
\section{Conclusion}
\vspace{-2mm}
This work develops a framework for knowledge distillation through a target-aware transformation that enables the student to aggregate the useful semantic over itself to enhance the expressivity of each pixel, which allows the student to act as a whole to mimic the teacher rather than minimize each partial divergence in parallel.
Our method is successfully extended to semantic segmentation by the proposed hierarchical distillation consisting of patch-group and anchor-point distillation, designed to focus on local feature and long-range dependency. We conduct thorough experiments to validate the effectiveness of the method and advance the state-of-the-art.

\vspace{-1mm}
\section{Discussion}
\vspace{-2mm}
\textbf{Potential negative societal impact.} Our method has no ethical risk on dataset usage and privacy violation as all the benchmarks are public and transparent.

\textbf{Limitations.} There are some issues of interest that we would like to explore in the future: (1) Currently, we only select the last layer of the backbone network for distillation. It would be interesting to see the efficacy when multiple layers are get involved with distillation which has been explored by some works \cite{Zagoruyko2017PayingMA,chen2021distilling}. (2) Also, we didn't investigate the effectiveness on other applications like object detection, which may need to design the new objective to fit the nature of specific application. 

\vspace{-3mm}
\section*{Acknowledgement}
\vspace{-2mm}
This work was supported in part by National Natural Science Foundation of China (NSFC) under Grant No.61976233, National Key Research and Development Program of China (Grant NO. 2020AAA0108104), Australian Research Council (ARC) Discovery Early Career Researcher Award (DECRA) under DE190100626, and Alibaba Innovative Research (AIR) Program.
\clearpage
{\small
\bibliographystyle{ieee_fullname}

}
\clearpage

\begin{figure*}
    \centering
    \includegraphics{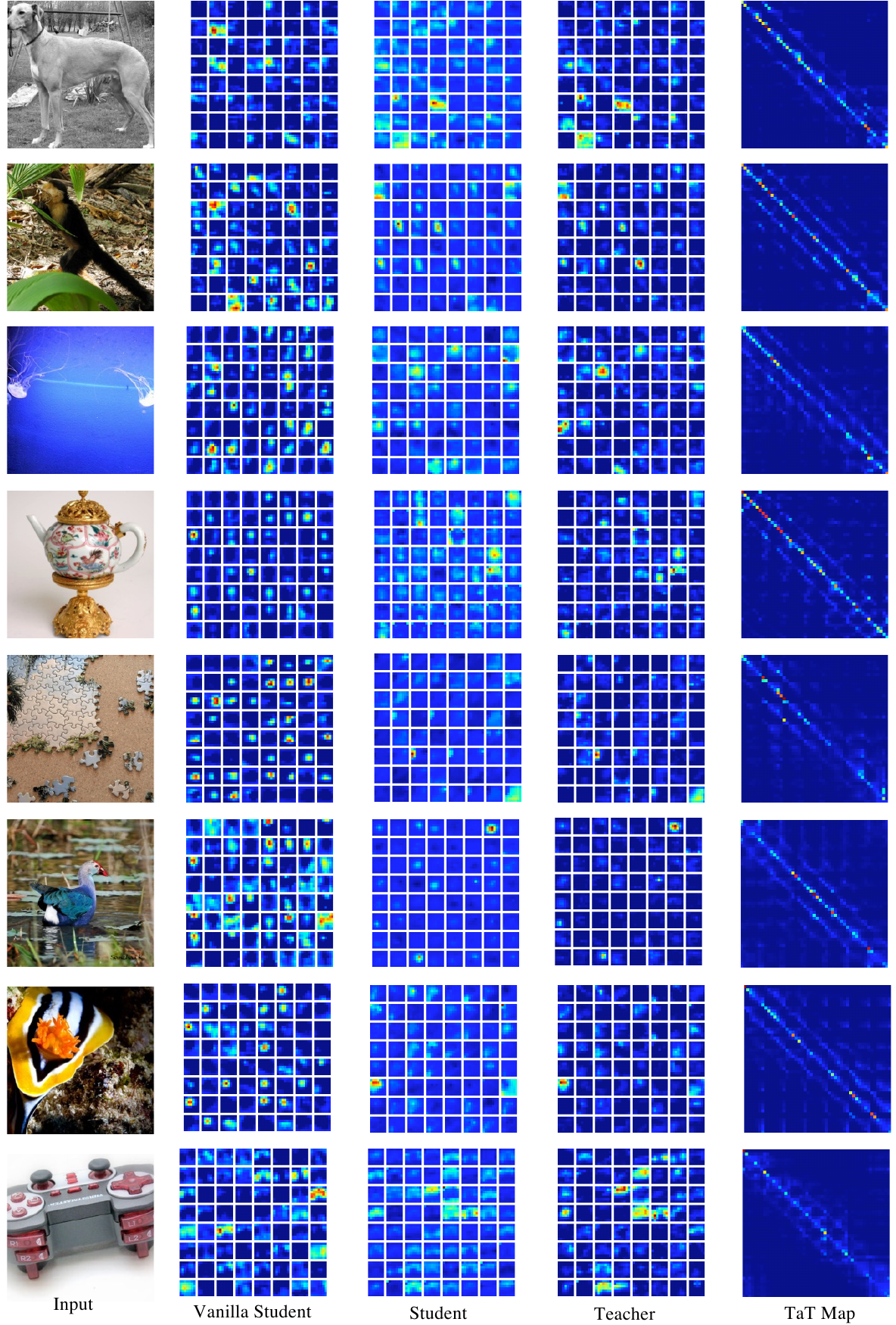}
    \vspace{-1mm}
    \caption{
    \textbf{Visualization of feature map and TaT map.} The input is selected from ImageNet validation set. The teacher backbone is ResNet34 and student backbone is ResNet18. The feature map of the distillation layer (4-th block) has been visualized. While there are 512 feature channels in total, we visualize 64 channels for better visualization. Through the Target-aware transformer, we found that the reconfigured student feature (3rd column) has a similar pattern with teacher feature (4th column). 
    The associated TaT map has also been visualized, which indicates the student would aggregate the semantic mostly from neighbor to enhance its pixels. 
    }
    \label{fig:supp_vis}
    \vspace{-4mm}
\end{figure*}

\twocolumn[
\begin{@twocolumnfalse}
\begin{center}
    
\huge Appendix
\end{center}
\section*{A.1 Asset Usage }
\label{sec:asset}
\ \quad This work is built upon some public dataset and code assets. We appreciate their efforts. The benchmark dataset has been introduced in main paper. Here we list the URL, version, and license of the code assets that we used:

\begin{center}
\captionof{table}{Usage of Code assets.}
\vspace{3mm}
\begin{tabular}{l|l|c|l}
    \toprule
     Exp.&\multicolumn{1}{c|}{URL} &Ver. &Licence \\\midrule
     ImageNet&{\tt \small https://github.com/yoshitomo-matsubara/torchdistill} &{\tt \small 7b883ec} &MIT\\ \midrule
     Cifar100&{\tt \small https://github.com/HobbitLong/RepDistiller}&{\tt \small 9b56e97} &BSD 2-Clause \\ \midrule
     \multirow{2}{*}{Pascal VOC}&{\tt \small https://github.com/jfzhang95/pytorch-deeplab-xception}&{\tt \small 9135e10} & MIT\\
     &{\tt \small https://github.com/clovaai/overhaul-distillation} &{\tt \small 76344a8} &MIT\\ \midrule
     \multirow{2}{*}{COCOStuff10k} &{\tt \small https://github.com/kazuto1011/deeplab-pytorch} &{\tt \small 4219467} &MIT\\
     &{\tt \small https://github.com/dvlab-research/ReviewKD} &{\tt \small cede6ea} &N/A\\
    \bottomrule
\end{tabular}
\label{tab:my_label}
\end{center}

\section*{A.2 Additional Experiments}
\subsection*{A.2.1 Comparison on COCOStuff10k}
\ \quad For the experiments of semantic segmentation, we have compared our method to a variety of stat-of-the-art methods in the Section 4 of the main paper. In terms of COCOStuff10k, since some methods do not support this dataset, we re-implement them and the result is presented on Table~\ref{tab:supp_seg}. We found that our method is competitive and it outperforms the comparison methods.

\begin{center}
\captionof{table}{Comparison (mIoU\%) on COCOStuff10k.}
\vspace{3mm}
\begin{tabular}{l|ccc}
\toprule
     &ICKD~\cite{Liu2021ICKD} &Overhaul~\cite{Heo2019ACO} &Ours  \\ \midrule
    ResNet18 &27.22 &27.86 &28.75 \\   \midrule
    MobileNetV2 &26.64 &26.96 &28.05 \\
\bottomrule
\end{tabular}
\label{tab:supp_seg}
\end{center}

\subsection*{A.2.2 Hyperparameters on Cifar-100}
\ \quad We used Bayesian optimization to obtain the weight factors $\alpha$ and $\epsilon$ in Eq. 9. Here we show the searching result on different backbones (See Table~\ref{tab:supp_cifar100_p}). We found that in most cases (4 out of 6), $\epsilon$ is greater than $\alpha$, which indicates that our proposed objective is more important than the standard Cross-entropy during distillation. For instance, in the distillation VGG13$\rightarrow$VGG8, $\epsilon$ is 8 and $\alpha$ is only 0.1. We also found that for the similar architectures, the searching result is similar, \eg, when WRN-40-2 and ResNet110 are selected as teacher.

\begin{center}
\captionof{table}{Coefficients $\alpha$ and $\epsilon$ on different backbones on Cifar-100.}
\vspace{3mm}
\begin{tabular}{c|ccccccc}
\toprule
    Teacher & WRN-40-2 & WRN-40-2  & ResNet56 & ResNet110 & ResNet110 & ResNet32$\times$4 & VGG13 \\
    Student & WRN-16-2 & WRN-40-1  & ResNet20 & ResNet20  & ResNet32  & ResNet8$\times$4  & VGG8 \\ \midrule 
    $\alpha$   &0.8 &0.7 &0.8 &1 &1 &6 &0.1\\
    $\epsilon$ &4 &3.6 &0.4 &0.75 &1 &39 &8\\
    \bottomrule
\end{tabular}
\label{tab:supp_cifar100_p}
\vspace{3mm}
\end{center}

\end{@twocolumnfalse}
]
\twocolumn[
\begin{@twocolumnfalse}
\begin{center}
\captionof{table}{Adding $\mathcal{L}_{\rm{KL}}$ on Cifar100.}
\vspace{0mm}
\resizebox{2.0\columnwidth}{!}{
\tablestyle{20pt}{1.0}
\begin{tabular}{l|cccc} 
\toprule
Teacher& WRN-40-2 &ResNet110 & ResNet32$\times$4 & VGG13 \\
Student& WRN-16-2 &ResNet20  & ResNet8$\times$4  & VGG8 \\ \midrule
KD &74.92 &70.67 &73.33 &72.98 \\
FitNet+KD &75.12 &70.67 &74.66 &73.22 \\
AT+KD &75.32 &70.97 &74.53 &73.48 \\
SP+KD &74.98 &71.02 &74.02 &73.49 \\
CC+KD &75.09 &70.88 &74.21 &73.04 \\
RKD+KD &74.89 &70.77 &73.79 &72.97 \\
PKT+KD &75.33 &70.72 &74.23 &73.25 \\
NST+KD &74.67 &71.01 &74.28 &73.33 \\
CRD+KD &75.64 &71.56 &75.46 &74.29 \\
ICKD+KD &75.57 &71.91 &75.48 &73.88 \\
Ours+KD &\textbf{76.08}&\textbf{72.16}&\textbf{75.54}&\textbf{74.35}\\
\bottomrule
\end{tabular}
}
\label{tab:cifar_add_kl}
\vspace{3mm}
\end{center}

\subsection*{A.2.3 Adding KD loss on Cifar-100}
\ \quad We report the result of our method in Table~\ref{tab:cifar_add_kl} with $\mathcal{L}_{\rm{KL}}$ loss to compare with the baselines under the same settings. Our method with KD loss surpasses all the baselines again. 

\subsection*{A.2.4 Feature Visualization}
\ \quad We further visualize the feature map and the associated TaT map to intuitively understand the functionality behind the proposed Target-aware Transformer. As exhibited in Figure~\ref{fig:supp_vis},  we visualize the feature maps of student before and after distillation, which are compared to the feature map of teacher. The teacher backbone is ResNet34 and student backbone is ResNet18. The input images are randomly selected from ImageNet validation set. While the 4-th block (\ie distillation layer) of ResNet34 and ResNet18 has 512 channels, we visualize 64 channels for better visualization. 

\ \quad Obviously, the reconfigured student feature (3rd column) has a more similar pattern with teacher feature (4th column), which demonstrates that TaT can effectively adapt the student to mimic the teacher. In terms of the TaT map, which controls the intensity of semantic aggregation, it is close to the identity matrix. Recall that we apply the linear function $\phi(\cdot)$ on student feature $f^s$.
And the TaT map will be further applied on $\phi(f^{s})$ to reconfigure the student feature, which is lately asked to minimize the L$_2$ distance with teacher feature. When the TaT map is an identity matrix, it means that $\phi(f^{s})$ can reconstruct the teacher feature on its own. However, since TaT map is not strictly the identity matrix, it indicates that each pixel of $\phi(f^{s})$ still needs to \textit{borrow} the semantic from other position (mostly neighborhood) to enhance itself. Indeed, by aggregating the semantic from neighbors, each pixel increases the receptive field and thus semantic capacity. This demonstrates the semantic mismatch between student and teacher due to the variation on network depth and width.
\end{@twocolumnfalse}
]

\end{document}